\documentclass[runningheads]{llncs}
\usepackage{graphicx}
\usepackage{amsmath,amssymb} 
\usepackage{color}
\usepackage{array}
\usepackage{multirow}
\usepackage{slashbox}
\newcommand{\minitab}[2][l]{\begin{tabular}{#1}#2\end{tabular}}
\usepackage[width=122mm,left=12mm,paperwidth=146mm,height=193mm,top=12mm,paperheight=217mm]{geometry}
\begin{document}
\pagestyle{headings}
\mainmatter

\title{Relay Backpropagation for Effective Learning of Deep Convolutional Neural Networks} 

\titlerunning{ }

\authorrunning{ }

\author{Li Shen$^{1}$, Zhouchen Lin$^{2}$, Qingming Huang$^{1}$}
\institute{$^{1}$ University of Chinese Academy of Sciences \\
$^{2}$ Key Lab. of Machine Perception (MOE), School of EECS, Peking University}

\maketitle

\begin{abstract}
Learning deeper convolutional neural networks becomes a tendency in recent years. However, many empirical evidences suggest that performance improvement cannot be gained by simply stacking more layers. In this paper, we consider the issue from an information theoretical perspective, and propose a novel method {\it Relay Backpropagation}, that encourages the propagation of  effective information through the network in training stage. By virtue of the method, {\it we achieved the first place in ILSVRC 2015 Scene Classification Challenge.} Extensive experiments on two challenging large scale datasets demonstrate the effectiveness of
our method is not restricted to a specific dataset or network architecture. Our models will be available to the research community later.

\keywords{Relay Backpropagation, Convolutional Neural Networks, Large scale image classification.}
\end{abstract}

\section{Introduction}

Convolutional neural networks (CNNs) are capable of inducing rich features from data, and have been successfully applied in a variety of computer vision tasks. Many breakthroughs obtained in recent years benefit from the advances of convolutional neural networks \cite{krizhevsky_nips2012,girshick_cvpr2014,taigman_cvpr2014,karpathy_cvpr2014}, spurring the research of pursuing a high performing network. The importance of network depth is revealed in these successes. For example, compared with AlexNet \cite{krizhevsky_nips2012}, the utilization of VGGNet \cite{simonyan_iclr2015} brings about substantial gains of accuracy on 1000-class ImageNet 2012 dataset by virtue of deeper architectures.

Increasing the depth of network becomes a promising way to enhance performance. On the downside, such solution is accompanied by the growth of parameter size and model complexity, that poses great challenges for optimization. The training of deeper networks typically encounters the risk of divergence or slower convergence, and prone to overfitting. Besides, many empirical evidences \cite{simonyan_iclr2015,Srivastava_icml2015,he_iccv2015} (e.g., the results reported by \cite{simonyan_iclr2015} on ImageNet dataset shown in Table~\ref{table:simple_stacking} (Left)) suggest that the improvement on accuracy cannot be trivially gained by simply adding more layers. It is in accordance with the results in our preliminary experiments on Places2 challenge dataset \cite{zhou_2015}, that deeper networks even suffer from a decline on performance (in Table~\ref{table:simple_stacking} (Right)).

\begin{table}[t]
\setlength{\abovecaptionskip}{10pt}
\setlength{\belowcaptionskip}{0pt}
\renewcommand\arraystretch{1.1}
\begin{minipage}{0.5\textwidth}
\begin{center}
\begin{tabular}{|p{1.8cm}|p{1.6cm}<{\centering}p{1.6cm}<{\centering}|}
\hline
\multirow{2}{*}{\minitab[l]{Model}} & \multicolumn{2}{c|}{ImageNet 2012} \\
& top-1 err. & top-5 err. \\
\hline
VGGNet-13 & 28.2 & 9.6 \\
\hline
VGGNet-16 & 26.6 & 8.6 \\
\hline
VGGNet-19 & 26.9 & 8.7 \\
\hline
\end{tabular}
\end{center}
\end{minipage}
\begin{minipage}{0.5\textwidth}
\begin{center}
\begin{tabular}{|p{1.8cm}|p{1.6cm}<{\centering}p{1.6cm}<{\centering}|}
\hline
\multirow{2}{*}{\minitab[l]{Model}} & \multicolumn{2}{c|}{Places2 challenge} \\
& top-1 err. & top-5 err. \\
\hline
VGGNet-19 & 48.5 & 17.1 \\
\hline
VGGNet-22 & 48.7 & 17.2 \\
\hline
VGGNet-25 & 48.9 & 17.4 \\
\hline
\end{tabular}
\end{center}
\end{minipage}
\caption{Error rates (\%) on ImageNet 2012 classification and Places2 challenge validation set.
The training and testing procedures for all models are the same.
VGGNet-22 and VGGNet-25 are obtained by simply adding 3 and 6 layers on VGGNet-19, respectively.}
\label{table:simple_stacking}
\end{table}

\begin{figure}
\setlength{\abovecaptionskip}{0pt}
\begin{center}
\includegraphics[width=\textwidth]{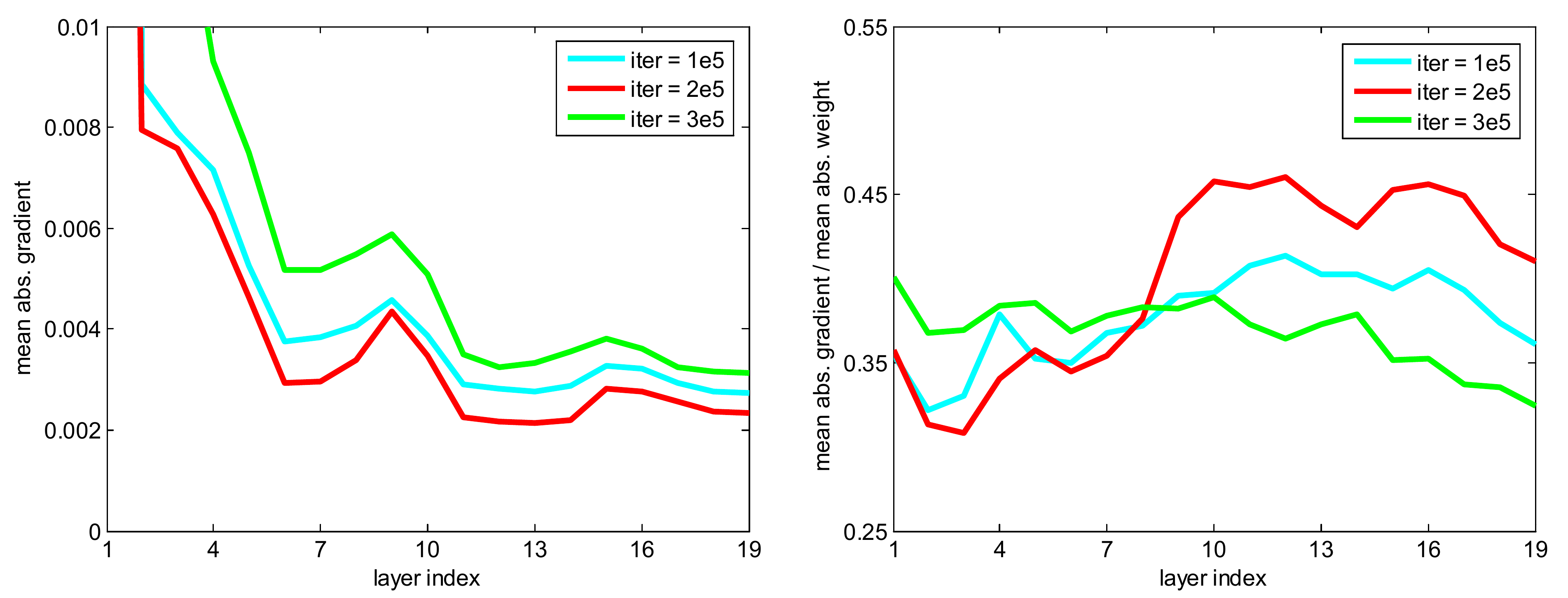}
\end{center}
\caption{Magnitude of the gradient at each convolutional layer of the 22-layer CNN model (i.e., 19 convolutional layers and 3 fully connected layers) in Table~1. A color line plot the gradient magnitude of different layers at a certain number of iterations. (Left) Average magnitude of gradients. (Right) Relative magnitude of gradients, i.e., average magnitude of gradients divided by average magnitude of weights.}
\label{fig:gradient_vanishing}
\end{figure}

To interpret the phenomenon, we should be concerned with the possibility of vanishing and exploding gradient which are the crucial reasons that hamper the training of very deep networks with backpropagation \cite{lecun_1998} (BP) algorithm, as gradients might be prone to either very small or very large after backpropagation with many layers. To investigate whether vanishing and exploding problems appear, we analyze the scale of the gradients at different convolutional layers during training. Take the 22-layer CNN model (in Table \ref{table:simple_stacking}) as an example, shown in Fig.~\ref{fig:gradient_vanishing}. The average magnitude of gradients and their relative values with respect to weights are displayed, respectively. We can observe that the gradient magnitude of lower layers does not tend to vanish or explode, but approximately remain stable when performing the optimization. In practice, the issues of vanishing and exploding gradient have been largely coped with by aid of some techniques, e.g., rectifier neuron \cite{nair_icml2010,maas_icml2013}, refined initialization scheme \cite{glorot_icais2010,he_iccv2015}, and Batch Normalization \cite{ioffe_2015}.

However, from an information theoretical perspective \cite{kaminura_2002,cover_2006,tishby_2015}, the amount of information about target outputs diminishes during propagation, although the gradient does not vanish. Such degradation will amplify as network goes deeper. In order to effectively update network parameters, the error information should not go back too many layers. However, the problem is inevitable when optimizing a very deep network with standard backpropagation algorithm.

\begin{figure}
\setlength{\abovecaptionskip}{0pt}
\begin{center}
\includegraphics[width=\textwidth]{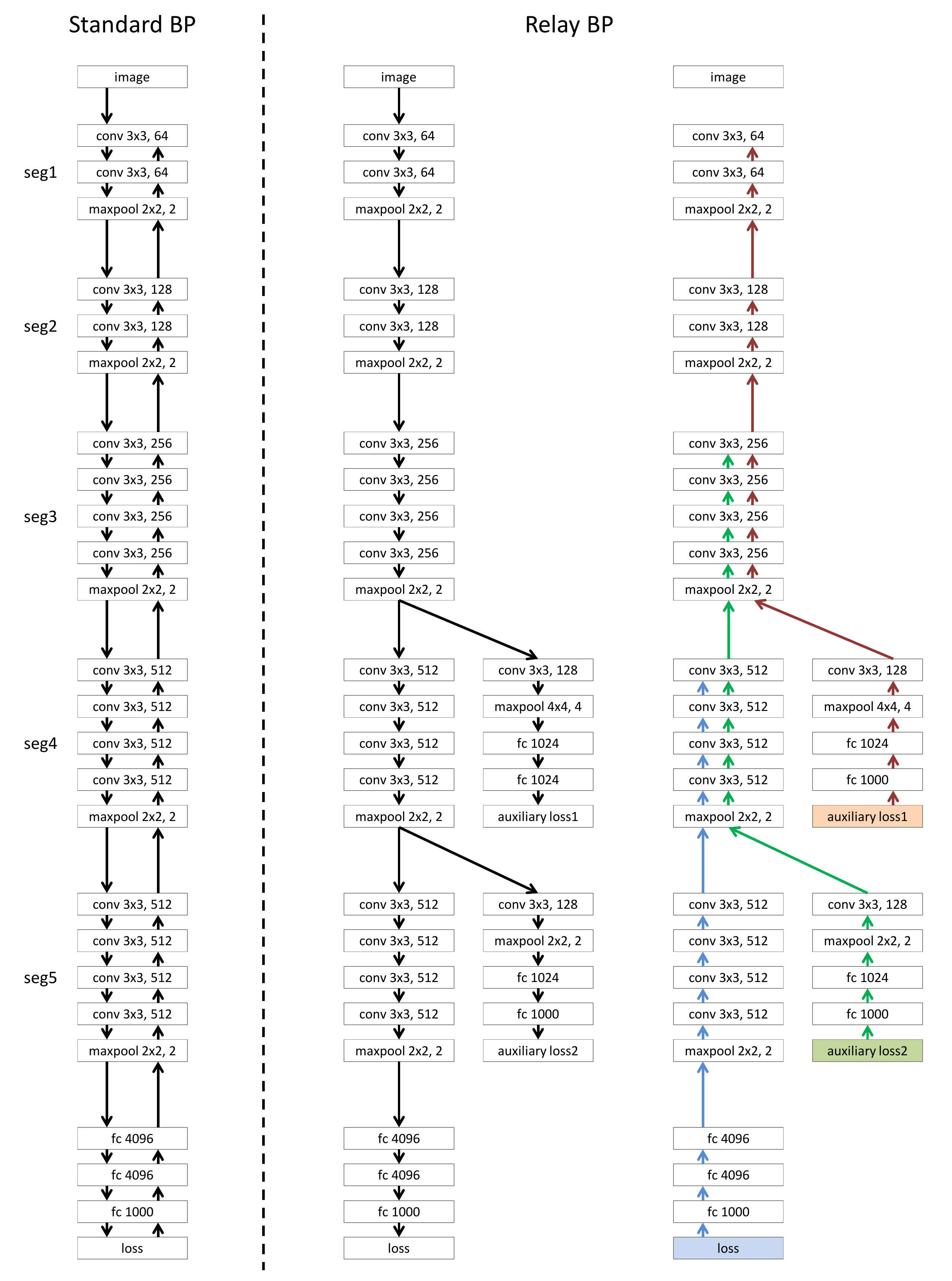}
\end{center}
\caption{(Left) VGGNet-19 network \cite{simonyan_iclr2015} with standard backpropagation algorithm. (Middle \& Right) VGGNet-19 extended network with Relay backpropagation algorithm. This is an example with two auxiliary output modules, adding two branches on traditional VGGNet-19 architecture. The black arrows denote the forward propagation of information through the network, and the color arrows indicate the information (gradient) flows at backward propagation. This figure is best viewed on the screen.}
\label{fig:relay_bp_framework}
\end{figure}

To address the problem, in this paper we propose a novel method, {\it Relay Backpropagation} (Relay BP), which encourages the propagation of {\it effective information} through the network in training stage. To accomplish the aim, the whole network is first divided into several segments. We introduce one or multiple interim output modules (including loss layer) after intermediate segments, and aim to minimize the ensemble of losses. More importantly, the gradients from different losses are propagated to the lowest layers of respective segments, namely, the gradient with respect to certain loss will propagate at most $N$ consecutive layers, where $N$ is smaller than realistic network depth. An example framework is depicted in Fig.~\ref{fig:relay_bp_framework} with two auxiliary output modules. In a word, we provide an elegant way to effectively preserve relevant information by shortening the path from outputs to lower layers, and meanwhile restrain the adverse effect of less relevant information which propagated through too many layers.

By virtue of Relay BP, we achieve the first place in ILSVRC 2015 Scene Classification Challenge, which provides a new large scale dataset involving 401 classes and more than 8 million training images. The benefits of the method are also verified on ImageNet 2012 classification dataset with another two famous network architectures, which demonstrates our method is not constrained to a specific architecture or dataset. We will make our models available to the research community.

\section{Related Work}

Convolutional neural networks have attracted much attention over the last few years.
For image classification tasks with large scale data \cite{zhou_nips2014,xiao_ijcv2014,russakovsky_ijcv2015}, there is a tendency of increasing the network complexity (e.g., the depth \cite{simonyan_iclr2015} and the width \cite{zeiler_eccv2014}), which brings about the difficulties of training the network. A range of techniques are exploited to address the problem by taking various angles. For example, Simonyan and Zisserman \cite{simonyan_iclr2015} propose to reduce the risk of divergence by initializing a deeper network with the aid of pre-training shallower ones. Refined initialization schemes are adopted to train very deep networks directly by drawing the weights from properly scaled distributions \cite{glorot_icais2010,he_iccv2015}. Moreover, the benefits of new activation functions \cite{nair_icml2010,he_iccv2015,goodfellow_2013} for training deep networks have been shown in extensive experiments. Besides, some studies are developed in the direction of finding better optimizers, such as stochastic gradient descent with momentum \cite{sutskever_icml2013} and RMSProp \cite{tieleman_2012}, which is widely used and works well in practice.

It is particularly worthy of comparing our method with the work in \cite{lee_2015,szegedy_cvpr2015}, where temporary branches including classifiers are attached to intermediate layers, and helps to propagate the supervision information to lower layers with shortcuts. However, such multi-loss mechanism neglects information reduction due to long-term propagation, and the adverse effect of less relevant information for lower layers. Different from it, our method can effectively preserve relevant information and meanwhile restrain the adverse effect of less relevant information, thus obtain a model with better performance. In \cite{szegedy_2015,he_cvpr2016}, powerful networks are obtained by adopting new structures, i.e., Inception module and Residual block, which are concurrent with our work, and also attend the ILSVRC 2015 Challenge. The two structures implement shortcut connections in different ways, however long-term propagation still exists when training a deeper network. Therefore, our contribution is orthogonal to these work, and network performance can be further improved benefitting from our method.

\section{Standard BP and Information Reduction}

When training a network (e.g., VGGNet-19) with standard BP algorithm, optimization at each iteration is comprised of forward propagation and backward propagation (as shown in Fig.~\ref{fig:relay_bp_framework} (Left)). The process of forward propagation is to feed the images into input layer, forward propagate through the network, and finally produce predicted variable in output layer. Then error is generated according to the difference with true (desired) values, and gradients with respect to loss are propagated backward through the network until the lowest layer (i.e., the first convolutional layer), which is backward propagation. The weights are updated according to respective gradients accordingly. The procedure suggests that error information received by lower layers has flowed through many intermediate layers, whose number arises along with the growth of network depth.

From an information theoretical point of view, a network defines the direction of information flow, which forms a Markov chain. Let $y$ denote the starting gradient (i.e., the gradient of loss with respect to the weight parameters of last full connected layer), which preciously represents the supervision signal, and $x$, $u$ respectively denote the gradients transmitted to the layers in turn, i.e., $y \to x \to u$. According to Data Processing Inequality \cite{cover_2006} $I(y;x) \ge I(y;u)$, the information contained in a signal is unable to increase during processing. In practice, the amount of information about the loss (i.e., supervised information) through the network is prone to reduction due to a series of complicated transformations, which means less relevant information obtained from loss is received by lower layers. Such effect will amplify when information propagates deeper, ultimately hamper the performance of the whole network. In order to effectively update network parameters, the information should not go back too many layers.

\section{Relay BackPropagation}

The motivation of our method is to propagate effective information through the network in backward propagation. We accomplish the target by using auxiliary output modules appropriately. Take VGGNet-19 network architecture for example, as shown in Fig.~\ref{fig:relay_bp_framework} (right). The whole network is first divided into several segments separated with max-pooling layers. For example, from the first convolutional layer to the first max-pooling layer is considered as a segment, and the next segment starts from the third convolutional layer and ends to the second max-pooling layer. Thus, there are totally five segments, numbered 1 to 5 from lower to higher layers.

We attach one or multiple auxiliary output modules to intermediate segments. Fig.~\ref{fig:relay_bp_framework} is an example with two output modules (i.e., auxiliary loss 1 and loss 2), which are added after segment 3 and 4, respectively. In order to preserve the relevant information about loss, the propagation of each loss is through at most $N$ consecutive layers, where $N$ is the upper limit of the numbers of layers that we deem that can carry enough relevant information. Namely, different losses are responsible for different parts of weight layers in the network. The information flows from different losses are represented with different colors in Fig.~\ref{fig:relay_bp_framework}. Auxiliary loss 1 (colored with red) would be propagated until the lowest one in segment 1, and auxiliary loss 2 (colored with green) would be propagated until the lowest one in segment 3, and the primary loss (colored with blue) would be propagated until the lowest layer in segment 4, respectively.

More importantly, there is overlapping between information flows at intermediate segments, such as segment 4 receives the information from primary loss and auxiliary loss 2. As our optimization objective is to minimize the sum of the three losses, the updating on segment 4 would fuse the information passed through from the two losses. Consequently, segment 4 plays the role of the transition between the two information flows of primary loss and auxiliary loss 2, not only the transition between lower and higher layers trivially, that is why we call the method as {\it Relay Backpropagation}. Such mechanism is capable of alleviating the optimization difficulty at lower layers and favoring better discrimination at higher layers, and achieving the aim of coordinating the information propagation in a very deep network ultimately.

In summary, there are two distinct characters in our method: (1) different losses are responsible for different parts of weight layers in the network, rather than the all layers below. This is different from traditional multi-loss with standard BP algorithm \cite{lee_2015,szegedy_cvpr2015}. Such mechanism is helpful to reduce the degradation of relevant information about loss and restrain the adverse effect of less relevant information due to long-term propagation. (2) information flows from different losses exist overlapping at intermediate segments, that guarantees coordinate information propagation in the network.

In forward propagation step, information transmission follows the manner from input to output layers, where the activations generated at one layer are fed into its adjacent layer in turn. The black arrows in Fig.~\ref{fig:relay_bp_framework} (middle) indicate the directions of information flows through the network. It is consistent with standard BP for the network with auxiliary branches.

When testing an image, a prediction is made without considering auxiliary branches, as auxiliary supervision is introduced only to enhance the training of network. Consequently, there is no extra cost (parameter size and time expense) brought in testing stage, ensuring the test efficiency of model.

One might be concerned with: Where to add auxiliary output module? And which segments (or convolutional layers) should belong to the scope of certain loss? We apply the heuristic scheme based on empirical evidences in this work. Nevertheless, some intuitive rules can be considered for guidance. One insight is that it is inadvisable to add auxiliary output modules at too lower layers, since the patterns captured at these layers lack of sufficient discrimination for recognizing a high-level concept (e.g., object or scene). Moreover, the depth of a network is an important factor to be considered. Adding an auxiliary branch might be enough if the network is not too deep. In general, the design can be adjusted flexibly according to specific requirements and practical experience.

\section{Experiments}
In this section, we evaluate Relay BP on Places2 challenge \cite{zhou_2015} and ImageNet 2012 classification dataset \cite{russakovsky_ijcv2015}, and also investigate it on four different network architectures. We show Relay BP outperforms baselines significantly. The baseline methods are briefly introduced below:

\begin{itemize}
\item \textbf{Standard BP: } Given the network, information forward and backward propagation follow the rule of traditional backpropagation algorithm (e.g., in Fig.~\ref{fig:relay_bp_framework}(Left)).
\item \textbf{Multi-loss + standard BP: } One auxiliary output module (branch) is attached to parts of intermediate layers.
\end{itemize}

For a fair comparison, the network architecture in training stage (i.e., the architecture with temporary branches) is identical for our method and the baseline of multi-loss with standard BP. The difference lies in the scheme of information backward propagation. In the experiments, we only add one auxiliary branch for all networks, as they are not too deep to tackle. Moreover, the increment of branch also brings about training computation cost. Therefore, the principle is adding the branches as few as possible. We intend to train extremely deeper networks by aid of multiple branches in future work.

\subsection{Places2 Challenge}
We evaluate our method on the Places2 challenge dataset \cite{zhou_2015}, which is used in ILSVRC 2015 Scene Classification Challenge. This dataset includes images belonging to 401 scene categories, with 8.1M images for training, 20K images for validation and 381K images for testing. To mimic the real-world frequencies of scene occurrence, there is a non-uniform distribution of images per category for training, ranging from 4,000 to 30,000. The classification performance of the challenge is evaluated using the top-5 error, which allows an algorithm to identify multiple scene categories for an image, because many environments can be described using different words.

\subsubsection{Network Architectures.} Relay BP is independent on the network architectures used. We investigate two types of deep convolutional neural network architectures on the Places2 challenge dataset, as shown in Table \ref{tab:architectures}.
The model A is based on VGGNet-19 \cite{simonyan_iclr2015}, and simply adds 3 convolutional layers on the three smaller feature maps (56, 28, 14).
The model B uses a $7\times7$ convolutional layers and a modified inception module as building block.
We also incorporate spatial pyramid pooling (spp) \cite{he_eccv2014} into the models, where the pyramid configuration is $7\times7$, $3\times3$, $2\times2$ and $1\times1$. Dropout regularization is applied to the first two fully-connected (fc) layers, with the dropout ratio 0.5.
We use Rectified Linear Unit (ReLU) as nonlinearity and do not use Batch Normalization \cite{ioffe_2015} in the two networks.
The experiments involving Batch Normalization will be seen in Section \ref{experiment:imagenet}.
The auxiliary classifier \normalsize\textcircled{\scriptsize{2}} is used in multi-loss standard BP and Relay BP, rather than standard BP.
The loss weight of the auxiliary classifier is set to 0.3.
The ``gradient'' in Table \ref{tab:architectures} shows the details of backward propagation in Relay BP.

\begin{table}[t]
\setlength{\belowcaptionskip}{0pt}
\renewcommand\arraystretch{1.1}
\begin{center}
\begin{tabular}{p{1.5cm}<{\centering}|p{1.5cm}<{\centering}|p{3.5cm}<{\centering}|p{5.0cm}<{\centering}}
\hline
input size & gradient & model A & model B \\
\hline
\multirow{2}{*}{\minitab[c]{224$\times$224}} & \multirow{2}{*}{\minitab[c]{\normalsize\textcircled{\scriptsize{2}}}}
& $[$\ conv 3$\times$3, 64\ $]$ $\times$\ 2 & $[$\ conv 7$\times$7, 128, stride 2\ $]$ $\times$\ 1 \\
& & maxpool 2$\times$2, 2 &  \\
\hline
\multirow{2}{*}{\minitab[c]{112$\times$112}} & \multirow{2}{*}{\minitab[c]{\normalsize\textcircled{\scriptsize{2}}}}
& $[$\ conv 3$\times$3, 128\ $]$ $\times$\ 2 &  \\
& & maxpool 2$\times$2, 2 & maxpool 2$\times$2, 2 \\
\hline
\multirow{2}{*}{\minitab[c]{56$\times$56}} & \multirow{2}{*}{\minitab[c]{\normalsize\textcircled{\scriptsize{2}}}}
& $[$\ conv 3$\times$3, 256\ $]$ $\times$\ 5 & $[$\ modified inception, k 64\ $]$ $\times$\ 4 \\
& & maxpool 2$\times$2, 2 & maxpool 2$\times$2, 2 \\
\hline
\multirow{2}{*}{\minitab[c]{28$\times$28}} & \multirow{2}{*}{\minitab[c]{\normalsize\textcircled{\scriptsize{1}}\normalsize\textcircled{\scriptsize{2}}}}
& $[$\ conv 3$\times$3, 512\ $]$ $\times$\ 5 & $[$\ modified inception, k 128\ $]$ $\times$\ 4 \\
& & maxpool 2$\times$2, 2 & maxpool 2$\times$2, 2 \\
\hline
\multirow{1}{*}{\minitab[c]{-}}
& - &\multicolumn{2}{c}{auxiliary classifier \normalsize\textcircled{\scriptsize{2}}} \\
\hline
\multirow{2}{*}{\minitab[c]{14$\times$14}} & \multirow{2}{*}{\minitab[c]{\normalsize\textcircled{\scriptsize{1}}}}
& $[$\ conv 3$\times$3, 256\ $]$ $\times$\ 5 & $[$\ modified inception, k 128\ $]$ $\times$\ 4 \\
& & spp, \{7, 3, 2, 1\} & spp, \{7, 3, 2, 1\} \\
\hline
- & \normalsize\textcircled{\scriptsize{1}} & \multicolumn{2}{c}{fc 4096} \\
\hline
- & \normalsize\textcircled{\scriptsize{1}} & \multicolumn{2}{c}{fc 4096} \\
\hline
- & \normalsize\textcircled{\scriptsize{1}} & \multicolumn{2}{c}{fc 401, classifier \normalsize\textcircled{\scriptsize{1}}} \\
\hline
\end{tabular}
\end{center}
\caption{Architectures of the networks used for ILSVRC 2015 Scene Classification.
The convolutional layer is denoted as ``conv $<$receptive field$>$, $<$filters$>$''.
The max-pooling layer is denoted as ``maxpool $<$region size$>$, $<$stride$>$''.
Our modified inception module concatenates the outputs of a 1$\times$1 convolution with k filters,
a 3$\times$3 convolution with k filters and two 3$\times$3 convolution with 2k filters.
\normalsize\textcircled{\scriptsize{1}} and \normalsize\textcircled{\scriptsize{2}}
indicate which layers the gradients propagate to.}
\label{tab:architectures}
\end{table}

\subsubsection{Class-aware Sampling.}
The Places2 challenge dataset has more than 8M training images in total. The numbers
of images in different classes are imbalanced, ranging from 4,000 to 30,000 per class.
The large scale data and non-uniform class distribution pose great challenges for model learning.

To address this issue, we apply a sampling strategy, named ``class-aware sampling'', during training.
We aim to fill a mini-batch as uniform as possible with respect to classes,
and prevent the same example and class from always appearing in a permanent order.
In practice, we use two types of lists, one is class list, and the
other is per-class image list. Taking Places2 challenge dataset for example, we have one class list, and 401 per-class image lists.
When getting a training mini-batch in an iteration,
we first sample a class X in the class list, then sample an image in
the per-class image list of class X. When reaching the end of the per-class image list of class X,
a shuffle operation is performed to reorder the images of class X.
When reaching the end of class list, a shuffle operation is performed to reorder the classes.
We leverage such a class-aware sampling strategy to effectively tackle the non-uniform class distribution,
and the gain of accuracy on the validation set is about 0.6\%.

\begin{table}[t]
\setlength{\belowcaptionskip}{10pt}
\renewcommand\arraystretch{1.1}
\begin{center}
\begin{tabular}{p{4.0cm}|p{1.8cm}<{\centering}p{1.8cm}<{\centering}|p{1.8cm}<{\centering}p{1.8cm}<{\centering}}
\hline
\multirow{2}{*}{\minitab[l]{\ Method}} & \multicolumn{2}{c|}{model A} & \multicolumn{2}{c}{model B} \\
& top-1 err. & top-5 err. & top-1 err. & top-5 err. \\
\hline
\ standard BP & $50.91$ & $19.00$ & $50.62$ & $18.69$ \\
\ multi-loss + standard BP & $50.72_{(0.19)}$ & $18.84_{(0.18)}$ & $50.59_{(0.03)}$ & $18.68_{(0.01)}$ \\
\ Relay BP & $49.75_{(1.16)}$ & $17.83_{(1.17)}$ & $49.77_{(0.85)}$ & $17.86_{(0.83)}$ \\
\hline
\end{tabular}
\end{center}
\caption{\textbf{Single crop} error rates (\%) on Places2 challenge validation set.
In the brackets are the improvements over ``standard BP'' baseline.}
\label{places2:single_crop}
\end{table}

\begin{table}[t]
\setlength{\belowcaptionskip}{0pt}
\renewcommand\arraystretch{1.1}
\begin{center}
\begin{tabular}{p{4.0cm}|p{1.8cm}<{\centering}p{1.8cm}<{\centering}|p{1.8cm}<{\centering}p{1.8cm}<{\centering}}
\hline
\multirow{2}{*}{\minitab[l]{\ Method}} & \multicolumn{2}{c|}{model A} & \multicolumn{2}{c}{model B} \\
& top-1 err. & top-5 err. & top-1 err. & top-5 err. \\
\hline
\ standard BP & $48.67$ & $17.19$ & $48.29$ & $16.89$ \\
\ multi-loss + standard BP & $48.55_{(0.12)}$ & $17.05_{(0.14)}$ & $48.27_{(0.02)}$ & $16.89_{(0.00)}$ \\
\ Relay BP & $47.86_{(0.81)}$ & $16.33_{(0.86)}$ & $47.72_{(0.57)}$ & $16.36_{(0.53)}$ \\
\hline
\end{tabular}
\end{center}
\caption{\textbf{Single model} error rates (\%) on Places2 challenge validation set.
In the brackets are the improvements over ``standard BP'' baseline.}
\label{places2:single_model}
\end{table}

\subsubsection{Training and Testing.} Our implementation is based on the publicly available code of Caffe \cite{jia2014_caffe}. We train models on the provided Places2 challenge training set, do not use any additional training data. The image is resized isotropically so that its shorter side is 256. To augment the training set, a 224$\times$224 crop is randomly sampled from a training image, with the per-pixel mean subtracted. The random horizontal flipping and standard color shift in \cite{krizhevsky_nips2012} are used. We initialize the weights using \cite{he_iccv2015} and train all networks from scratch. We train the networks by applying stochastic gradient descent (SGD) with mini-batch size of 256 and a fixed momentum of $0.9$. The learning rate is initialized to 0.01, and is annealed by a factor of 10 when the error plateaus.
The training is regularized by weight decay (set to 0.0002). We train all models up to $80\times10^4$ iterations. In testing, we take the standard ``single crop (center crop)" protocol in \cite{szegedy_cvpr2015}. Furthermore, we use the fully-convolutional testing \cite{simonyan_iclr2015} to report the performance of single model. The image is resized isotropically so that its shorter side is in \{224, 256, 320, 384, 448\}, and the scores are averaged at multiple scales.

\subsubsection{Comparisons of Results.} Table~\ref{places2:single_crop} lists the results of the three methods with ``single crop" testing strategy. Compared with standard BP, the baseline ``Multi-loss + standard BP" shows better performance by introducing auxiliary supervision on intermediate layers, however the superiority is marginal, even negligible with regard to model B. In contrast, our method achieves significant improvement over standard BP, as well as consistently outperforms ``Multi-loss + standard BP" (approximately $1.0\%$ on model A and $0.8\%$ on model B based on top-5 measure). It is notable that the improvement on model B is less than the one on model A. The shortcut connections in modified Inception modules make it possible to propagate information with shortcuts, somewhat alleviates the information reduction. This is also the reason of ineffectiveness of ``Multi-loss + standard BP" on model B. Nevertheless, our method is capable of improving the performance on model B. It confirms our insight that restraining the adverse effect of less relevant information is helpful for training deep neural networks.

For a comprehensive comparison, we also report the model performance with ``single model" testing strategy in Table~\ref{places2:single_model}. Clear advantage can be observed in our method compared to the baselines. It is worthy of mentioning that the improvement of single model over center crop is less, about $1.5\%$ top-5 error diminished from $17.83\%$ (single crop) to $16.33\%$, while empirical results on ImageNet 2012 classification dataset suggest the performance gain is approximately $3.0\%$ \cite{he_iccv2015,ioffe_2015}.

\subsubsection{ILSVRC 2015 Scene Classification Challenge.}

\begin{table}[t]
\setlength{\belowcaptionskip}{0pt}
\renewcommand\arraystretch{1.1}
\begin{center}
\begin{tabular}{p{4.5cm}|p{2.4cm}<{\centering}}
\hline
\ Team name & top-5 err. \\
\hline
\ Ntu\_rose & 19.33 \\
\ Trimps-Soushen & 17.98 \\
\ Qualcomm Research & 17.59 \\
\ SIAT\_MMLAB & 17.36 \\
\hline
\ WM (model A) & 17.35 \\
\ WM (model B) & 17.28 \\
\ \textbf{WM (model ensemble)} & \textbf{16.87} \\
\hline
\end{tabular}
\end{center}
\caption{The competition results of ILSVRC 2015 Scene Classification.
The top-5 error rates (\%) is on Places2 challenge test set and reported by the test server.
Our submissions are denoted as ``WM''.}
\label{places2:ILSVRC}
\end{table}

\begin{figure}[t]
\setlength{\abovecaptionskip}{0pt}
\begin{center}
\includegraphics[width=\textwidth]{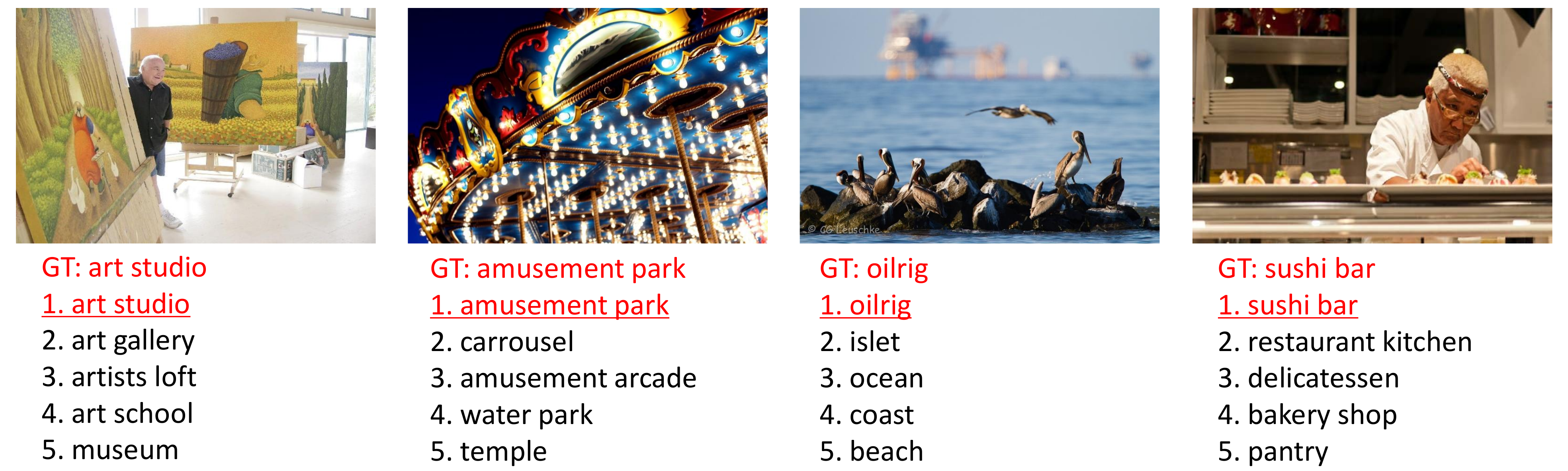}
\end{center}
\caption{Example images successfully classified by our method on Places2 challenge validation set.
For each image, the ground-truth label and our top-5 predictions are listed.}
\label{fig:successfully}
\end{figure}

\begin{figure}[t]
\setlength{\abovecaptionskip}{0pt}
\begin{center}
\includegraphics[width=\textwidth]{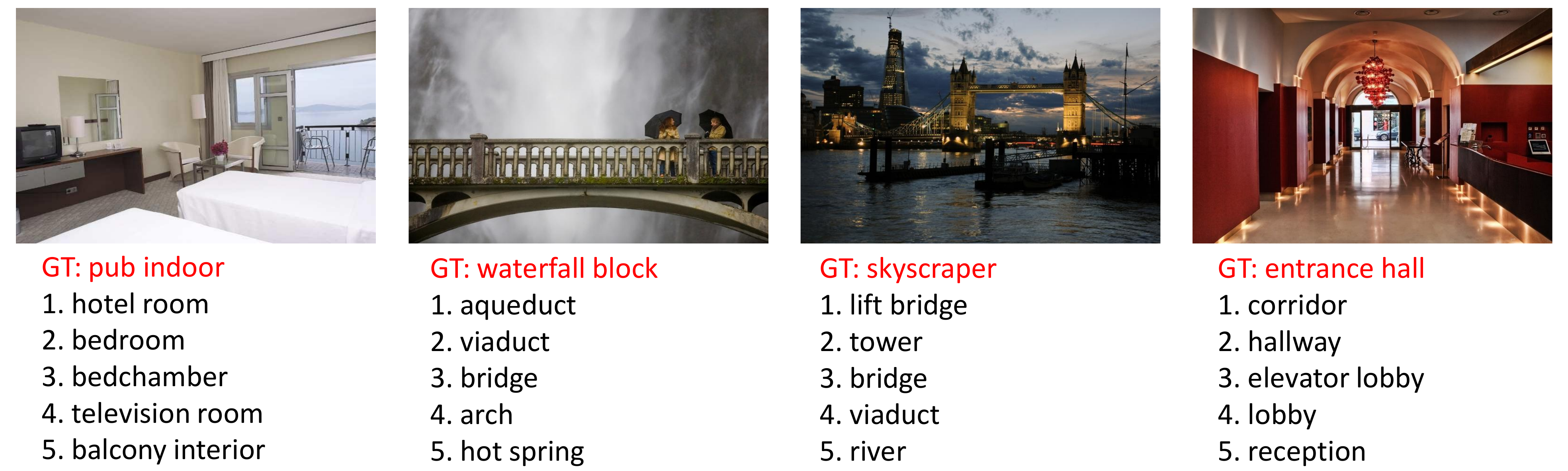}
\end{center}
\caption{Example images incorrectly classified by our method on Places2 challenge validation set.
For each image, the ground-truth label and our top-5 predictions are listed.}
\label{fig:incorrectly}
\end{figure}

{\it By virtue of Relay BP, our ``WM'' team won the 1st place in ILSVRC 2015 Scene Classification task.} Table \ref{places2:ILSVRC} shows the results of this challenge. Our five entries won the top five places, and the results of our single model A and B outperform all ensemble results from other teams. We combine five models of different architectures and input scales, and achieve 15.74\% top-5 error on validation set. Our top-5 error is 16.87\% on the testing set, which is roughly 1.1\% worse than the validation result. We conjecture that there might be a distribution gap between validation and testing data, because similar degradation has also been observed by other teams \cite{zhou_2015}. The improvement of model ensemble over single model B is 0.4\%. From single crop to single model, and further to model ensemble, the improvement is consistently lower than expected. We conjecture that training with large scale data enhances the capability of single view, leading to the difficulties of further improvement with model ensemble.

Fig.~\ref{fig:successfully} shows some example images of Places2 challenge validation set, which are successfully classified by our method.
The predicted labels are in descending order of confidence. Even though many high-level scene concepts exist large intra-class appearance variance, our method can still recognize them easily. On the other hand, we also show some incorrectly classified examples in Fig.~\ref{fig:incorrectly}. These predictions seem to be reasonable, although they fail in context of evaluation measure.
A scene image might be typically described by multi-labels. Moreover, the composing of a scene is mostly complicated, such as a place is comprised of multiple objects and the same object might appear in different places. The loose connections between scene and object concepts increase the difficulties of scene recognition.

\subsection{ImageNet 2012 Classification}
\label{experiment:imagenet}
We evaluate our method on the ImageNet 2012 classification dataset \cite{russakovsky_ijcv2015}, which has become one of the benchmarks to assess the progress of image classification. This dataset includes images belonging to 1000 classes, with 1.2M images for training, 50K images for validation and 100K images for testing. The classification performance is measured by the top-1 and top-5 error rates. We use the provided data for training models, do not use any additional data.

\subsubsection{Configurations.} Recently, residual networks \cite{he_cvpr2016} introduce shortcut connections, and has achieved state-of-the-art performance on ImageNet 2012 classification dataset. Moreover, \cite{szegedy_2015} utilizes the "Inception-v3" architectures, and yielded comparable classification accuracy. We use the 50-layer residual network (ResNet-50) \cite{he_cvpr2016} and the Inception-v3 architectures \cite{szegedy_2015} to evaluate Relay BP. For both architectures, we do not use scale jitter augmentation \cite{simonyan_iclr2015} during training. Standard SGD is applied to train the networks. Other configurations (including data augmentation, network architectures, and training/testing methodology) remain unchanged as \cite{he_cvpr2016,szegedy_2015}.
More details about the configurations can be found in \cite{he_cvpr2016,szegedy_2015}. For Relay BP, we add one auxiliary branch with the loss weight set to 0.3. The gradient overlapping segments of primary and auxiliary loss range from ``conv4\_1'' to ``conv4\_4'' (ResNet-50), and ``inception4a'' to ``inception4d'' (Inception-v3), respectively. As the scheme of multi-loss has been included in Inception-v3, we omit the baseline ``Multi-loss + standard BP" in Table~\ref{imagenet:single_model}.

\subsubsection{Results Analysis and Discussion.}

\begin{table}[t]
\setlength{\belowcaptionskip}{0pt}
\renewcommand\arraystretch{1.1}
\begin{center}
\begin{tabular}{p{4.5cm}|p{1.4cm}<{\centering}|p{1.4cm}<{\centering}p{1.4cm}<{\centering}|p{1.4cm}<{\centering}p{1.4cm}<{\centering}}
\hline
\multirow{2}{*}{\minitab[l]{\ Method}} & \multirow{2}{*}{\minitab[l]{dataset}} & \multicolumn{2}{c|}{ResNet-50} & \multicolumn{2}{c}{Inception-v3} \\
& & top-1 err. & top-5 err. & top-1 err. & top-5 err. \\
\hline
\ standard BP \cite{he_cvpr2016,szegedy_2015} & val & 20.74 & 5.25 & 18.77 & 4.20 \\
\hline
\ standard BP (re-implement) & \multirow{2}{*}{\minitab[l]{val}} & 21.17 & 5.37 & 19.18 & 4.43 \\
\ Relay BP & & 20.26 & 4.93 & 18.52 & 3.97 \\
\hline
\ Relay BP & test & - & 4.95 & - & \textbf{4.03} \\
\hline
\end{tabular}
\end{center}
\caption{\textbf{Single model} error rates (\%) on ImageNet 2012 classification dataset.}
\label{imagenet:single_model}
\end{table}

Table~\ref{imagenet:single_model} lists the classification errors achieved in single model. The results in the first row are the ones reported in \cite{he_cvpr2016} and \cite{szegedy_2015}, respectively. And the second row displays the results by our re-implementation. There is slight difference between the two rows, mainly because of the diversity of details in implementation, which has been described in the section of "Configurations".

The models trained with Relay BP achieve better classification performance compared to the ones trained with standard BP. The accuracy improvement is $0.44\%$ on top-5 measure, and $0.91\%$ on top-1 measure based on ResNet-50 network. Besides, there are $0.46\%$ and $0.66\%$ improvement on top-5 and top-1 measure based on Inception-v3 architecture. The common characteristic of the two architectures is the utilization of shortcut connections, although the implementations are different. As we have mentioned in above sections, shortcuts make the gradient of final outputs easily reach lower layers, thus are able to prevent the information reduction due to long-term propagation. This is also the evidence of only adding one auxiliary branch in Relay BP. Nevertheless, the network performance can be enhanced by aid of our method, which further demonstrates the promise of our insight that restraining the adverse effect of less relevance information is effective for improving network performance. Because of the high baselines, the improvement is so difficult, which highlights the effectiveness of our method. Moreover, we also report the results on test dataset (submitted to test server) to verify that the obtained results are not overfitting to the dataset. We only submitted the two results in the last half year, and the result $4.03\%$ outperforms the best result of single model reported in ILSVRC 2015 ImageNet Classification task.

\section{Conclusion}
In this paper, we proposed the method {\it Relay Backpropagation}, which encourages the transmission of effective information in backward propagation of training deep convolutional neural networks. Relevant information can be effectively preserved, and the adverse effect of less relevant information can be restrained. The experiments with four different network architectures on two challenging large scale datasets demonstrate the effectiveness of our method is not restricted to certain network architecture or specific dataset. As a future direction, we are interested in theoretical and mathematical support for the method.

\bibliographystyle{splncs}
\bibliography{egbib}
\end{document}